# Robustness, Evolvability and Phenotypic Complexity: Insights from Evolving Digital Circuits

**Nicola Milano, Paolo Pagliuca and Stefano Nolfi**
Institute of Cognitive Sciences and Technologies,
National Research Council (CNR),
Via S. Martino della Battaglia, 44
00185 Roma, Italia

Abstract
We show how the characteristics of the evolutionary algorithm influence the evolvability of candidate solutions, i.e. the propensity of evolving individuals to generate better solutions as a result of genetic variation. More specifically, (1+λ) evolutionary strategies largely outperform (μ+1) evolutionary strategies in the context of the evolution of digital circuits --- a domain characterized by a high level of neutrality. This difference is due to the fact that the competition for robustness to mutations among the circuits evolved with (μ+1) evolutionary strategies leads to the selection of phenotypically simple but low evolvable circuits. These circuits achieve robustness by minimizing the number of functional genes rather than by relying on redundancy or degeneracy to buffer the effects of mutations. The analysis of these factors enabled us to design a new evolutionary algorithm, named Parallel Stochastic Hill Climber (PSHC), which outperforms the other two methods considered.

## 1. Introduction

Robustness and evolvability are fundamental properties of biological systems. For the purpose of this paper we focus on robustness to mutations defined as the capability of a system to preserve its functionalities after mutations. Moreover, we define evolvability as the propensity of a system to produce adaptive heritable phenotypic variations as a result of mutations. These terms do not have an unique definition, for other usage see: Houle 1992; Wagner and Altenberg, 1996; Kirschner and Gerhart 1998; Bedau & Packard 2003; Earl and Deem 2004; Sniegowski and Murphy 2006; Wagner 2008; Masel and Trotter 2010.

To exemplify the nature of robustness and evolvability let us compare biological organisms with human designed computer programs (Wagner and Altenberg, 1996). Both are encoded in strings of characters: DNA in organisms, binary code in programs. However, they widely differ in terms of robustness and evolvability. Indeed, the former tend to preserve their function and to adapt as a result of random mutations of their sequence. The latter instead tend to completely loose their functionalities and have literally zero probability to improve as a result of random mutations.

At a first sight, robustness and evolvability have an antagonistic relationship: the higher the robustness of a system, the lower the probability that it will vary as a result of mutations is, and, consequently, the lower the evolvability of the system will be. Indeed, mechanisms that prevent changes such as proofreading and DNA repair enhance robustness but reduce evolvability (Lenski, Barrick and Ofria, 2006; Masel and Trotter, 2010). On the other hand, robustness to mutations facilitates the retention of mutations that enable the population to spread over large areas of the genetic space (Wagner, 2008). This combined with the fact that the phenotypes located on distant regions of the genetic space are much more varied than the phenotypes located in nearby regions of the genetic space increases the number of different phenotypes that can be produced, at the level of the population, through genetic variations (Wagner, 2008).

However, the relationship between robustness and evolvability can be influenced also by whether robustness is achieved through the utilization of phenotypically simple solutions that minimize the number of genes playing a functional role (de Visser et. al. 2003) or through phenotypically more complex solutions capable of buffering the effect of mutations through redundancy (i.e. through the utilization of multiple redundant components playing the same function) or degeneracy (i.e. through the utilization of multiple components playing multiple functions, see Tononi, Sporns, and Edelman, 1999; Edelman and Gally, 2001). And the way in which robustness is achieved can be influenced by the characteristics of the evolutionary process, in particular, by the level of competition among evolving individuals.

To investigate this hypothesis we measured robustness and evolvability in digital circuits evolved with different evolutionary methods. We chose digital circuits since they have been widely used in artificial evolutionary studies (Koza, 1992; Thompson Layzell and Zebulum 1999; Miller et al., 2000) and since they share with natural systems (e.g. proteins, RNA, regulatory circuits and metabolic networks) the following properties (Wagner, 2011; Raman and Wagner, 2011): (i) any phenotype (i.e. any circuit computing a given logic function) can originate from many different genotypes, (ii) these genotypes, giving rise to the same phenotype, can vary significantly among themselves, (iii) these genotypes span over vast genotype networks or neutral networks (Raman and Wagner, 2011: Shuster et al., 1994), i.e. genotypes giving rise to the same phenotype connected through single locus variation links, (iv) genotypes typically have many neighbors with the same phenotype and are thus robust to some extent to mutations, (v) the neighborhood of genotypes belonging to the same neutral network includes genotypes that give rise to rather different phenotypes. Moreover, we chose this domain since it permits to define in an operational and measurable way the key properties that we want to analyze, namely robustness, phenotypic complexity, evolvability, and phenotypic variability (i.e. the propensity of a system to produce adaptive heritable phenotypic variations as a result of mutations, independently of whether or not variations are adaptive). A more detailed definition of these terms and the description of the way in which these properties are measured in the context of digital circuits are reported in section 2.3.

The obtained results indicate that the level of competition between individuals subjected to differential reproduction influences the phenotypic complexity and the evolvability of the individuals. The strength of the effect is remarkable as demonstrated by the fact that circuits evolved with the $(1 + \lambda)$ evolutionary strategy (Rechenberg, 1973) solve the problem relatively quickly in all replications while circuits evolved with the $(\mu + 1)$ evolutionary strategy solve the problem in only 57% of the replications.

The analysis of the evolutionary dynamics also provided the basis for the development of a new original algorithm, called parallel stochastic hill climber (PSHC), which permits to achieve better results.

## 2. Method

In this section we describe the digital circuits, the evolutionary algorithms, and the measures used to analyze the evolutionary dynamics.

### 2.1 The digital circuits

Digital circuits (Figure 1) are systems that compute digital logic functions, such as the multiplication of digital numbers, by receiving as input two or more binary (Boolean) values and by producing as output one or more binary values. They are composed of multiple logic gates that receive in input two binary values (from the input pattern and/or from the output of other logic gates) and produce in output one binary value by computing an elementary logic function (OR, AND, NAND ext.) of the input. The logic function computed by a circuit depends on the functions computed by its constituent logic gates and by the way in which they are wired.

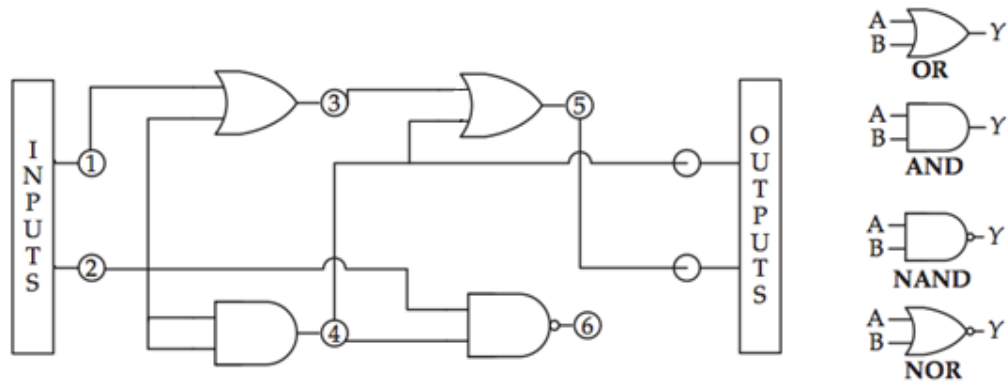

Figure 1. A digital circuit with two inputs, two outputs, and four gates. The right side of the panel shows the four symbols that correspond to the four kinds of permissible logic gates. The numbers 1-2 indicate the binary states that are provided as inputs to the circuit (input pattern). The numbers 3-6 indicate the output computed by the four corresponding logic gates. The output of the circuit corresponds to the output of the two logic gates that are wired to the output units (gate 4 and 5 in the case of this example). The lines indicate the way in which gates are wired.

Digital circuits can be realized in hardware or simulated in a computer. In standard electronic digital circuits the number and type of gates and the way in which they are wired is hardwired and hand-designed. In reconfigurable electronic digital circuits (such as the FPGA, see Balch, 2003), instead, the logic function computed by each gate and the way in which gates are wired can be varied. In evolvable hardware applications or more generally in evolutionary circuits the logic functions computed by each gates and the way in which gates are wired are encoded in artificial genotypes and evolved. Evolving circuits are selected on the basis of their fitness that is usually computed by measuring the extent to which the function computed by a circuit approximates a given target function (Thomson, Layzell and Zebulum, 1999).

In our experiments we evolved simulated digital circuits with four inputs, 400 logic gates divided into 20 layers of 20 gates, and one output for the ability to compute a 5-bit even parity function (i.e. to produce as output 1 when there is an even number of 1s in the input pattern and 0 otherwise). We choose this function since it constitutes a rather difficult problem for evolving circuits including OR, AND, NAND, and NOR logic gates (Miller and Thomson, 2000).

As in several related works (Thompson, Layzell and Zebulum, 1999; Hartmann and Haddow, 2004; Raman and Wagner, 2011), we choose to provide digital circuits with a fixed number of gates since this enable us to use a simple encoding schema. For alternative approaches that enable to evolve circuits having a variable number of gates see for example Miller and Hartmann (2001) and Macia and Solé (2009). Notice however that the utilization of a fixed number of logic gates only limit the maximum size of the circuits. Indeed, as we will see, evolving circuits typically rely on a much smaller number of gates with respect to the limit, i.e. they include several non-functional gates that do not contribute to the function computed by the circuit itself (see for example the gate computing the state 6 in Figure 1). In other words, the functional size of the evolving circuits can vary freely, within the upper limit imposed by the maximum number of gates. This also implies that the number of genes encoding phenotypical components (gates) playing a functional role can also vary freely during evolution.

Circuits are evaluated for the ability to map the $2^n$ possible input patterns into the corresponding desired outputs (i.e. 1 for input patterns with an even number of 1 and 0 otherwise). The fitness of the circuits is calculated on the basis of the following equation:

$$F = 1 - \frac{1}{2^n} \sum_{j=1}^{2^n} |O_j - E_j| \qquad (1)$$

where *n* is the number of inputs of the circuit, *j* is the number of the input pattern varying in the range [1, $2^n$], $O_j$ is the output of the circuit for pattern *j*, $E_j$ is the desired output for pattern *j*.

The genotype of evolving circuit is constituted by a vector of integer numbers that encodes the type of function computed by each logic gates and the way in which gates are wired. This approach has been named Cartesian Genetic Programming (Miller and Thomson, 2000). More specifically each genotype includes 400*3 genes that specify the type and the inputs of each logic gate and 1 additional gene that specifies the ID number of the gate that determines the output of the entire circuit. The genes that encode the characteristic of each logic gate include three integers that represent the type of the logic gate and the ID number of the inputs of the gate. The values of the genes are bounded in the range [1,4] in the case of integers encoding the type of the logic gate (1=OR, 2=AND, 3=NAND, 4=NOR), in the range [1, 5+(L-1)*20] in the case of the integers that encode the ID of the inputs of logic gates (where L is the layer of the corresponding logic gate), and in the range [6, 406] in the case of the gene that encodes the ID of the gate that constitutes the output of the entire circuit. Mutations are realized by replacing each integer with a certain probability (*MutRate*) with a number randomly generated with a uniform distribution in the appropriate range.

We will use the term behavior to indicate the outputs produced by a circuit in response to each possible input pattern. Moreover we will use the term functional size to indicate the number of gates that actively contribute to the outputs produced by the circuit. Notice that circuits having the same fitness might differ at the level of the behavior produced. Indeed, circuits producing different outputs can produce the same number of correct and incorrect responses. Notice also that circuits displaying the same behavior might differ at the level of the circuit's components. Indeed, circuits characterized by different type of gates or different wiring can produce the same outputs. The fact that the number of circuits that differ with respect to the type of the gates and/or the way in which the gates are wired is much greater than the number of circuits that differ at the level of the fitness implies that a large portion of genetic variations are neutral, i.e. produce variations at the level of the circuit and/or at the level of behavior that do not alter the fitness of the circuit.

## 2.2 Evolutionary algorithms

To investigate the relation between the characteristics of the algorithm and the evolvability of the circuits we compared three different algorithms.

The first is a $(1 + \lambda)$ evolutionary strategy (ES) that operates on the basis of a single parent, produces $\lambda$ offspring, and selects the best between the parent and the offspring as a new parent (see the pseudo-code below). We choose this algorithm since it turned out to be very effective in the evolution of digital circuits (Miller, Job and Vassiley, 2000).

To regulate the effect of the selection pressure we added to the fitness of the evolving individuals a value randomly selected in the range [-*Stochasticity*, *Stochasticity*] with a uniform distribution (see Jin, 2005). We decided to use this technique in combination with a selection operator that always selects the best between parents and offspring, rather than probabilistic selection operators such as roulette wheels or tournament selection, since: (i) it enables to regulate the selective pressure quantitatively by varying a single parameter (see Back, 1994), (ii) it enables to regulate the selective pressure from the maximum value, in which only the best candidate solutions between parents and offspring are selected, to a minimal value, in which the differential reproductive probability of better and worse individuals is minimal (see Back, 1994), (iii) it is qualitatively similar to the stochastic variation of fitness caused by uncontrolled variations occurring in uncertain environmental conditions. An example of uncertain conditions is constituted by the experiments involving robots in which the behavior displayed by the agent and consequently the fitness measure tend to vary during repeated evaluations even in apparently identical conditions as a consequence of small differences in the initial position and orientation of the robot, lighting conditions etc. Previous demonstrations of the fact that the introduction of noise in the fitness measure promotes the evolution of better solutions are reported in (Bäck and Hammel, 1994; Levitan and Kauffman, 1994; Rana, Whitlev, and Cogswell, 1996).

**Algorithm 1.** $(1 + \lambda)$ ES

1: genotype[0].generateRandomly()
2: genotye[0].fitness = genotype[0].evaluate()
3: **while** Evaluations < MaxEvaluations
4:     genotye[0].noisefitness= genotype[0].fitness + rand(-Stochasticity, Stochasticity)
5:     **for** o in range(1, $\lambda$+1)
6:         genotype[o] = genotype[0]
7:         genotype[o].mutate()
8:         genotye[o].fitness = genotype[o].evaluate()
9:         genotype[o].noisefitness[o] = genotype[o].fitness + rand(-Stochasticity, Stochasticity)
10:    **end for**
11:    genotype.sort(key=noisefitness, order=descending)
12: **end while**

The second algorithm is a $(\mu + 1)$ ES that operates with $\mu$ parents, enables each parent to produce a single offspring, and select the $\mu$ best individuals between the parents and the offspring as new parents (see the pseudo-code below). This type of algorithm is widely used for the evolution of neural networks (e.g. Harvey, 2001; Nolfi, et al., 2016). Also in this case we use the *Stochasticity* parameter to regulate the selective pressure.

Notice that in this method the offspring of individuals that are more robust to mutations have more chances to be selected than the offspring of individuals that are less robust. To illustrate this point consider the case in which two parents have the same fitness and in which the former parent is more robust to mutations than the second parent. Moreover, imagine that the mutations received by offspring are neutral or counter-adaptive (none of the mutations produce an improvement of the fitness). The offspring of the former parent have a greater probability to preserve the fitness of the parent than the offspring of the latter parent. Consequently, the offspring of the parent that is more robust has a greater probability to be selected. Overall this implies that the competition between the evolving individuals is regulated primarily by their fitness and secondarily (in the absence of adaptive mutations) by their robustness.

The competition between the individuals evolved with the $(1 + \lambda)$ method, instead, is not influenced by robustness to mutations. This is due to the fact that offspring originate from the same parent. Offspring might differ among themselves in term or robustness as a result of the mutations that they received. However, mutations that alter the level of robustness without altering the fitness do not provide an adaptive advantage in the $(1 + \lambda)$ method since two offspring with the same fitness have exactly the same probability to be selected, irrespectively of whether the former is more robust of the latter or vice versa.

**Algorithm 2.** $(\mu + 1)$ ES

```
1: for p in range(0, μ)
2:    genotype[p].generateRandomly()
3:    genotype[p].fitness = genotype[p].evaluate()
4: end for
5: while Evaluations < MaxEvaluations
6:    for o in range(0, μ):
7:       genotype[o].noisefitness = genotype[o].fitness += rand(-Stochasticity, Stochasticity)
8:       genotype[o + μ] = genotype[o]
9:       genotype[o + μ].mutate()
10:      genotype[o + μ].fitness = genotype[o + μ].evaluate()
11:      genotype[o + μ].noisefitness = genotype[o + μ].fitness
12:                                      += rand(-Stochasticity, Stochasticity)
13:   end for
11:   genotype.sort(key=noisefitness, order=descending)
18: end while
```

The third algorithm designed by us called Parallel Stochastic Hill Climber (PSHC) consists of a combination of a (μ + 1) and a (1 + 1) ES (see pseudo-code below). In this algorithm each parent is adapted through an (1 + 1) ES for a certain number of *Variations* during which the parent or a varied version of the parent generates a single mutated candidate solution that is discarded or used to replace the previous candidate solution depending on whether or not it is outperformed by the original solution. The best candidate solution obtained during this variation phase is then used to replace the parent and, with a certain low probability (*Interbreeding*), the worst individual of the population. The combined usage of the two types of evolutionary strategies enables to combine the advantages of operating on a population of parents with the advantages that can be gained by reducing the level of competition between the members of the population.

Also in the case of this algorithm we regulate the selection pressure during the variation phase by adding to the fitness of the candidate solutions a value randomly selected in the range [-*Stochasticity*, *Stochasticity*] with a uniform distribution. However, the selection of the best candidate solution obtained during the variation phase is made on the basis of the actual fitness (i.e. the fitness without noise). Moreover, whether or not the best variation is used to replace the original candidate solution and (eventually) another parent individual is based on the actual fitness (i.e. the fitness without noise). This technique should enable to maximize the advantage of stochasticity while eliminating the risk to fail retaining the best candidate solutions discovered to date.

**Algorithm 3. Parallel Stochastic Hill Climber (PSHC)**

```
1: for p in range(0, λ)
2:    genotype[p].generateRandomly()
3:    genotype[p].fitness = genotype[p].evaluate()
4: end for
5: genotype.sort(key=fitness, order=descending)
6: while Evaluations < MaxEvaluations
7:    for p in range (0, λ)
8:       var-genotype[0] = genotype[p]
9:       for v in range(1, NVariations)
10:         var-genotype[v] = var-genotype[v-1]
11:         var-genotype[v].mutate()
12:         var-genotype[v].fitness = var-genotype[v].evaluate()
13:         if (var-genotype[v].fitness + rand(-Stochasticity, Stochasticity)
14:            >= var-genotype[v-1].fitness + rand(-Stochasticity, Stochasticity))
15:            do nothing
```

```
16:         else
17:            var-genotype[v] = var-genotype[v-1]
18:            var-genotype[v].fitness = var-genotype[v-1].fitness
19:         endif
20:      end for
21:      var-genotype.sort(key=fitness, order=descending)
22:      if (var-genotype[0].fitness >= genotype[p].fitness)
23:           genotype[p] = var-genotype[0]
24:           genotype[p].fitness = var-genotype[0].fitness
25:      endif
26:   end for
27:   genotype.sort(key=fitness, order=descending)
28:   if (rand(0.0, 1.0) < Interbreeding and var.genotype[0].fitness > genotype[λ − 1].fitness)
29:        genotype[λ − 1] = var-genotype[0]
30:        genotype[λ − 1].fitness = var-genotype[0].fitness
31:   endif
32: end while
```

The PSHC is a form of island evolutionary algorithm (Whitley, Rana, and Heckendorn, 1998) in which the population is divided into a number of sub-populations and in which the individuals of each population are enabled to compete with the individuals of the other sub-populations only occasionally. In the case of the PSHC, however, the islands are constituted by single individuals that adapt on the basis of an $(1 + \lambda)$ evolutionary strategy instead than by sub-populations in which individuals compete with the other members of the sub-population. This difference is important since, as pointed out above, the competition between evolving individuals influences the robustness and the evolvability of the agents. Indeed, experiments carried out by using a variation of the algorithm in which the islands were formed by 2, 5, or 10 individuals led to significantly worse results (see below).

The evolutionary process is continued until a maximum number of total evaluations are performed. This number was set to 6 millions. Given that the computational cost of the selection and reproduction process is negligible with respect to the cost of the evaluation of candidate solutions, this enables to maintain the computational cost of experiments carried with different algorithms approximately constant. Each experiment is replicated 30 times with randomly different initial populations.

To verify that the differences in performance are not influenced by the parameters setting we replicated the experiments by systematically varying the parameters and we compared the results achieved by different algorithms in the experiments performed with the optimal parameters.

**2.3 Measures**

*Robustness* to mutations, defined as the capability of circuits to preserve their functionality after mutations, is measured in two way: (i) by calculating the fraction of offspring that have a fitness equal or greater than the fitness of the parent (fraction of fitness preserving offspring), and (ii) by calculating the fraction of offspring generated by performing a single random mutation that have a fitness equal or greater than the fitness of the parent. To estimate robustness with good precision the measures are calculated over 10,000 variations of the parent's circuit. The first measure is influenced by the mutation rate that might vary in different experiments while the second measure is independent from it.

*Phenotypic complexity*, defined as the complexity of the functional elements that constitute the system and the complexity of the way in which the elements are organized, is measured by counting the number of gates that contribute to generate the output of the circuit (functional gates). These gates are identified by marking as functional the gate connected to the output of the circuit, and then recursively the gates connected to gates connected to the output of the circuit. Given that in our experimental setup genes encode the logic function performed by each gate and the way in which the gate is wired to the other gates, this measure also indicates the number of genes that encode functional properties of the system. The term phenotypic complexity does not have a unique definition (see for example see Adami, 2002; Carlson and Doyle, 2002; Crutchfield and Görnerup, 2006; Hazen, et al., 2007; Whitacre, 2010). Moreover, phenotypic complexity is hard to define formally and to measure in the case of systems composed by heterogeneous elements and/or displaying multi-level organizations. The possibility to use a relatively straightforward measure in the case of our experiments, therefore, is due to the utilization of simple feed-forward digital circuits composed by homogeneous elements.

*Phenotypic variability*, defined as the propensity of circuits to vary phenotypically as a result of mutations, is measured (following Raman and Wagner [2011]) by calculating the number of unique functions computed by circuits located in the genetic neighborhood of the original circuit. This number is estimated performing for 10 times a 1,000 steps function-preserving random walk from the original circuit. The random walk is realized by: (i) generating a mutated circuit, (ii) incrementing a counter if the varied circuit computes a function that differ from the functions computed by the original circuit and by previous varied circuits, (iii) preserving or removing the mutation depending on whether the varied circuit has the same fitness of the original circuit or not, (iv) repeating the previous three operations for 1000 steps. Raman and Wagner (2011) refer to this measure with the term evolvability rather than with the term phenotypic variability. We prefer to distinguish between the propensity of a circuit to vary phenotypically independently of whether the variations are adaptive or nor (phenotypic variability) and the propensity of the circuit to improve its fitness (evolvability).

*Evolvability,* defined as the propensity of circuits to increase their fitness as a result of mutations*,* can also be measured by counting the number of new unique functions, fitter than the original circuit, computed by circuits located in the genetic neighborhood of the original circuit. This measure, however, depends on the relative fitness of the original circuit, i.e. the higher the fitness of the original circuit, the lower the number of circuits computing better functions is. Moreover, it is hard to estimate with sufficient precision given that the fraction of circuits computing better functions is small. For these reasons it is usually better measure evolvability indirectly by calculating the fraction of evolutionary experiments that find optimal solutions or the average number of evaluations required to find optimal solutions.

**2.4 Replicability**

Readers might replicate all the experiments described in this paper by downloading and installing FARSA from "https://sourceforge.net/projects/farsa/" and the required experimental plugin and configurations files from http://laral.istc.cnr.it/res/complexevolv/EvenParity.zip.

**3. Results**

In this section we describe the results obtained and the analysis performed.

**3.1 Digital circuits evolved with the ($\mu + 1$) ES**

In Table 1 we report the results of the experiment carried out by evolving digital circuits with the ($\mu + 1$) ES. To analyze the role of the critical parameters, we carried out 32 series of experiments in which we systematically varied the mutation rate and the stochasticity (Table 1). $\mu$ was set to 20.

|  | MutRate 0.01 | MutRate 0.02 | MutRate 0.03 | MutRate 0.04 |
|---|---|---|---|---|
| Stochasticity 0.00 | 46.66 (37.80) | 43.33 (33.91) | 33.33 (30.73) | 40.00 (32.65) |
| Stochasticity 0.01 | 46.66 (39.11) | 53.33 (35.83) | 36.66 (34.01) | 33.00 (33.41) |
| Stochasticity 0.02 | 30.00 (37.25) | 46.66 (37.53) | 43.33 (35.46) | 30.00 (32.48) |
| Stochasticity 0.03 | 43.33 (39.89) | 33.33 (38.25) | 53.33 (38.40) | 40.00 (35.62) |
| Stochasticity 0.04 | 30.00 (39.34) | 50.00 (36.24) | 30.00 (37.78) | 40.00 (38.74) |
| Stochasticity 0.05 | 50.00 (41.81) | **56.66** (42.87) | 53.33 (38.59) | 46.66 (37.95) |
| Stochasticity 0.06 | 43.33 (40.91) | 43.33 (36.69) | 50.00 (33.61) | 33.33 (33.11) |
| Stochasticity 0.07 | 30.00 (35.94) | 33.33 (34.34) | 30.00 (38.63) | 23.33 (37.23) |

Table 1. Performance of circuits evolved with the $(\mu + 1)$ ES in experiments carried out by using different mutation rate and stochasticity levels. $\mu$ was set to 20. The first number of each cell indicates the fraction of replications that achieved optimal performance. The numbers between parentheses indicate the average size of the evolved circuits.

As can be seen, in the case of the best combination of parameters (MutRate=0.02, Stochasticity=0.05), evolving circuits find optimal solutions in 56.66% of the replications. The best performance is achieved with an intermediate level of mutation rate (0.02) and an intermediate level of Stochasticity (0.05), see Table 1.

The analysis of evolved circuits indicates that, overall, they are very robust with respect to mutations. Indeed, 35.12% of the offspring of the circuits evolved with the optimal parameters have a fitness equal or greater than their parent and 95.15% of the mutations are neutral (i.e. do not alter the fitness). This high robustness, however, is largely due to the fact that the size of the functional part of the evolving circuits is small, i.e. only about 10% of the 400 gates actively contribute to the generation of the outputs of the circuits (see Table 1). Indeed, by restricting the analysis to the mutations that affect the functional gates, the percentage of neutral mutations drops from 95.15% to 0.52%.

As shown in Figure 2, the phenotypical complexity of the circuits, defined as the number of gates that contribute to generate the outputs of the circuits, increases during the initial phase of the evolutionary process (during which the fitness of the evolving circuits increases) and decreases during the successive phase (in which the fitness of the evolving circuits remains stable). This is the result of combined effect of the complexification of the individuals' phenotype driven by adaptive variations and the simplification of the individuals' phenotype driven by neutral variations. The complexification originates as a consequence of the strong correlation between the fitness and the functional size of the circuits (Figure 3, Spearman Test, rho 0.6359, phi $< 10^{-20}$ n = 600). The selection circuits fitter than their ancestors, in fact, leads to the selection of circuits which are phenotypically more complex than their ancestors, on the average. The simplification of the individuals' phenotype, instead, is caused by two factors. First, the fraction of neutral variations that produce a reduction of the number of functional gates is larger than the fraction of neutral variations that produce an increase of the number of functional gates (Table 2). Secondly, as stated above, the competition between reproducing individuals leads to the selection of individuals that are robust with respect to genetic variations. Given the strong negative correlation between the robustness and the phenotypic complexity of individuals (Figure 4, Spearman Test, rho -0.9017, phi $< 10^{-216}$ n = 600), in the majority of the cases the selection of robust individuals leads to the selection of phenotypically simpler individuals. The fact that the phenotypic complexity increases during the initial phase of the evolutionary process in which the fitness of the evolving individuals also increases and then decreases during the successive phase is thus due to the fact that the complexification factor operates during the initial evolutionary phase only while the simplification factors operates during the entire evolutionary process and especially during neutral evolutionary phases.

| larger | equal | smaller |
|---|---|---|

| 1.3820% | 97.0299% | 1.5881% |

Table 2. Probability to generate offspring with a larger, equal, or smaller number of functional gates than the parents. Data obtained by analyzing the offspring of 600 circuits evolved with the $(\mu + 1)$ ES in 30 replications carried out with the best parameters. Data computed by allowing each circuit to generate 10,000 offspring. The probability to generate smaller circuits is significantly greater than the probability to generate larger circuits (Wilcoxon test, p-value < $10^{-9}$).

The presence of a strong positive correlation between the size and the phenotypic variability of the circuits (Fig. 5, Spearman Test, rho 0.8897, phi < $10^{-204}$, n = 600) implies that the tendency to select phenotypically simple circuits leads to the selection of low evolvable individuals. In other words, the competition between reproducing individuals drives the evolving population toward a highly robust but low evolvable area of the search space. This area corresponds to the central and most connected region of the neutral network in which the fraction of mutations that does not alter the fitness of the circuits is maximum (Wilke, 2001) but in which the fraction of mutations that give rise to different unique phenotypes is minimum.

A correlation between the size and the phenotypic variability of circuits has already been reported by Raman and Wagner (2011). In their case the correlation was observed by comparing randomly generated circuits of different size that computed the same logic function. The correlation, therefore, seems to characterize all circuits, irrespectively of whether or not they were evolved and irrespectively of the function computed.

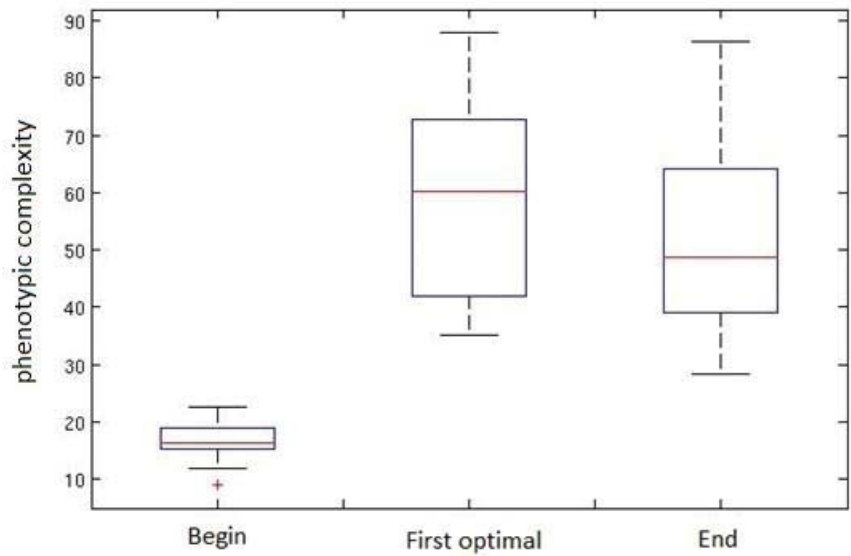

Figure 2. Phenotypic complexity of circuits evolved with the (μ+1) ES. The box-plots indicate the phenotypic complexity of best circuits of the first generation ("Begin"), of the first circuits that achieved optimal performance ("First optimal"), and of the circuits of the last generation ("End"). Data obtained by analyzing the circuits evolved in the 17 replications carried out with the best parameters that achieved optimal performance.

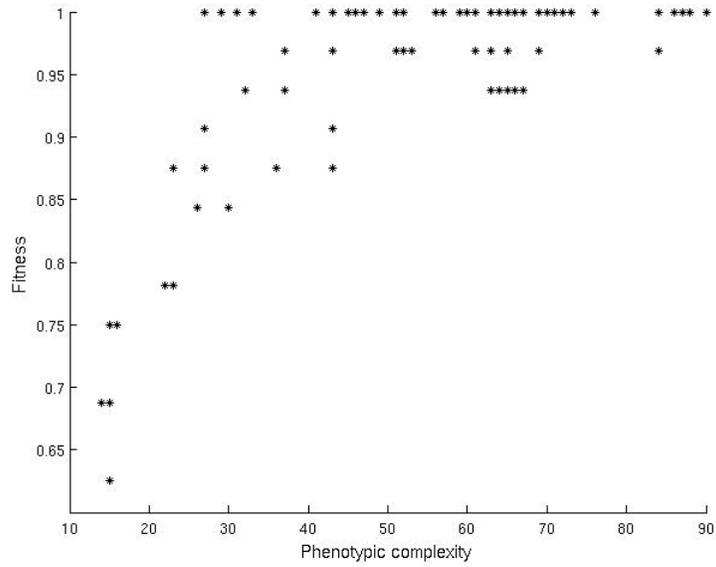

Figure 3. Scatter plot of fitness against phenotypic complexity in circuits evolved with the (μ+1) ES in 30 replications carried with the best parameters.

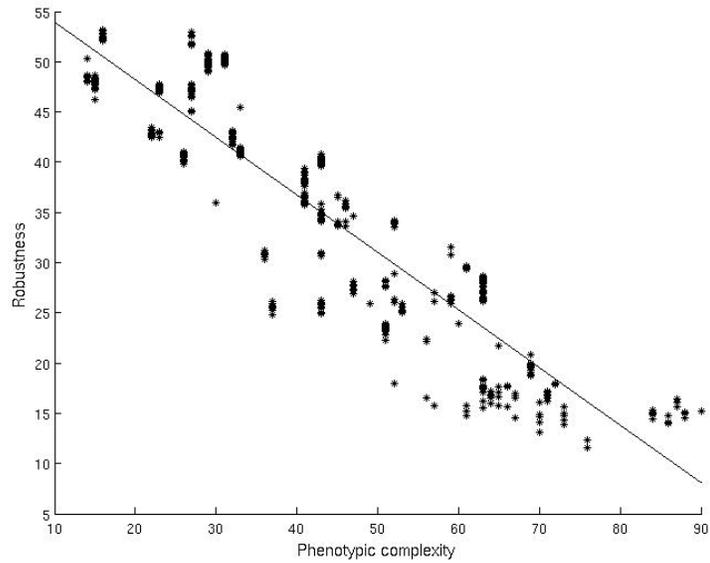

Figure 4. Scatter plot of robustness (i.e. fraction of offspring that maintain the same fitness of their parents) and phenotypic complexity in circuits evolved with the (μ+1) ES in 30 replications carried with the best parameters.

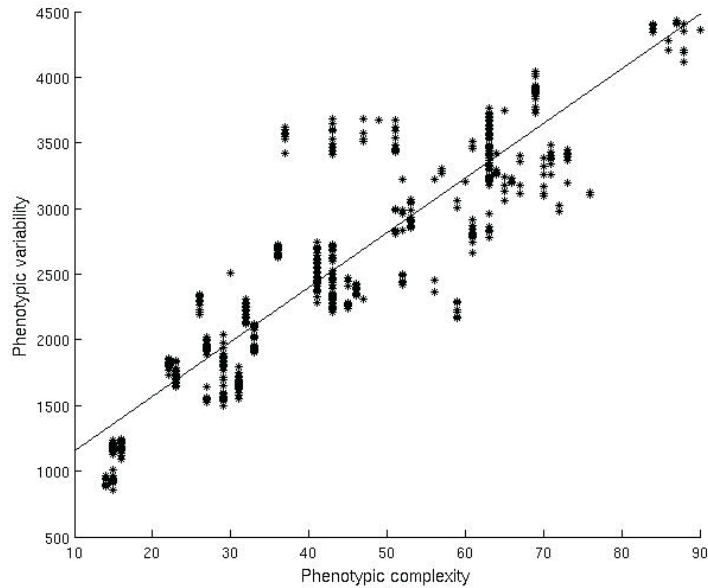

Figure 5. Scatter plot of phenotypic variability against phenotypic complexity in circuits evolved with the (µ+1) ES in 30 replications carried with the best parameters.

## 3.2 Digital circuits evolved with the $(1 + \lambda)$ ES

The analysis of the results obtained with the $(1 + \lambda)$ ES (Table 3 and 4) indicates that circuits evolved with this method are much more evolvable than the circuits evolved with the $(\mu + 1)$ ES. Indeed, the circuits evolved with the $(1 + \lambda)$ ES find optimal solutions in all replication after 301,400 evaluations, on the average. This performance is significantly better than the performance of the circuits evolved with the $(\mu + 1)$ ES (Wilcoxon rank sum p-value < $10^{-121}$).

These results are in line with those reported in previous studies (Miller, Job, and Vassiley, 2000) and confirm the efficacy of the $(1 + \lambda)$ ES in this domain. Notice also how, in the case of this method, the optimal results are achieved without Stochasticity, see Table 3). The experiments reported in Miller, Job, and Vassiley (2000) were also carried out without Stochasticity (the method used did not include the possibility to reduce the selection pressure through the addition of noise to the fitness).

|  | MutRate 0.01 | MutRate 0.02 | MutRate 0.03 | MutRate 0.04 |
|---|---|---|---|---|
| Stochasticity 0.00 | 96.66 (43.84) | *[384600] (42.38) | ***[301400] (45.50)** | *[395400] (35.58) |
| Stochasticity 0.02 | *[856070] (48.91) | *[544900] (42.76) | *[549700] (42.49) | *[463300] (40.54) |
| Stochasticity 0.04 | 86.66 (49.53) | *[541800] (43.11) | 96.66 (39.96) | 96.66 (40.71) |
| Stochasticity 0.06 | 43.33 (34.21) | 30 (26.37) | 3.33 (26.52) | 0 (25.77) |

Table 3. Performance of circuits evolved with the $(1 + \lambda)$ ES in experiments carried out by using different mutation rate and stochasticity levels. $\lambda$ was set to 10. The first number of each cell indicates the fraction of replications that achieved optimal performance. The experiments in which all replications found optimal solutions are indicated with an asterisk followed by the average number of evaluations that were necessary to find optimal solutions indicated with square brackets. The numbers between parentheses indicate the average size of the evolved circuits.

| N. Offspring ($\lambda$) |  |
|---|---|
| 5 | *[408400] (44.54) |
| 10 | ***[301400] (45.50)** |

| | | |
|---|---|---|
| 20 | *[524900] (43.85) | |

Table 4. Performance of circuits evolved with the $(1 + \lambda)$ ES in experiments carried out by using different number of offspring $(\lambda)$. The mutation rate and stochasticity parameters were set to 0.03 and 0.0, respectively. All experiments found optimal solutions. The numbers indicated with square brackets represent the average number of evaluations that were necessary to find optimal solutions. The numbers between parentheses indicate the average size of the evolved circuits.

At this point we should try to explain why the $(1 + \lambda)$ ES largely outperforms the $(\mu + 1)$ ES.

The variability of the population is null in the case of $(1 + \lambda)$ ES. Consequently, the superiority of this method cannot be due to the diversity of the population. The advantage is rather explained by the fact that this method leads to the selection of circuits that are less robust but that have a greater phenotypic variability than the circuits evolved with the $(\mu + 1)$ ES (Table 5).

The lower performance of the circuits evolved with the $(\mu + 1)$ ES, on the other hand, is due to the fact that it drives the population toward a very robust but low evolvable region of the genetic space. As we pointed out above, the competition between evolving circuits, in the experiments carried out with the $(\mu + 1)$ ES, drives the population toward solutions that are robust to genetic variations. Since the easiest way to achieve robustness consists in selecting solutions that are phenotypically simple and which have a low phenotypic variability, the selection of solutions that are robust to mutations produce a reduction of phenotypic variability that causes evolutionary stagnation.

The fact that the $(1 + \lambda)$ ES is able to find optimal solutions in all replications without stochasticity (see Table 3) indicates that the stagnation phases affecting the circuits evolved with the $(\mu + 1)$ ES is not caused only by the characteristics of the fitness surface (i.e. by the presence of local minima) but rather by the combined effect of the fitness surface and the tendency of the population to move toward areas of the genetic space containing highly robust but phenotypically simple solutions.

We can schematize this process by considering all the candidate solutions with the same fitness as a series of nodes and the genetic variations that transform parents into offspring with the same fitness as links between the nodes. The nodes and the links form one or more neutral networks (Shuster et al., 1994; Van Nimwegen, Crutchfield and Huynen, 1999) constituted by connected candidate solutions. Robustness is higher in the central part of the network in which the number of connections between nodes is maximum and lower in peripheral parts of the network. On the other hand phenotypic variability is lower in the central part of the network and is higher in the periphery (see also Hu et al., 2012). Stagnation thus originates from the tendency of circuits evolved with the $(\mu + 1)$ ES to move toward the central portion of the neutral network that has a low phenotypic variability.

| | $(\mu + 1)$ | $(1 + \lambda)$ | Wilcoxon p-value |
|---|---|---|---|
| Phenotypic complexity (number of functional gates) | 61.0 | 61.8 | $> 0.05$ |
| Phenotypic variability | 3826.3 | 4826.0 | $< 10^{-5}$ |
| Robustness (% offspring with a fitness equal or greater than the parent) | 23.7565 | 14.3 | $< 10^{-7}$ |

Table 5. Comparison of the characteristics of the first circuits evolved with the $(\mu + 1)$ and the $(1 + \lambda)$ ES that achieved optimal performance (i.e. the 17 out of 30 replications that achieved optimal performance in the case of the $(\mu + 1)$ experiments and 30 out of 30 replications in the case of the $(1 + \lambda)$ experiments).

**3.3 Digital circuits evolved with the PSHC algorithm**

Table 6 shows the results of the experiment carried out by evolving circuits with the parallel stochastic hill climber (PSHC) algorithm. To analyze the role of the critical parameters, we carried out 20 series of experiments in which we systematically varied the mutation rate and the stochasticity. The Variations parameter was set to 100 and the size of the population was set to 20. As can be see the best results are obtained without stochasticity and with the mutation rate parameter set to 0.02. Table 7 shows the results obtained in 7 additional series of experiments in which the Variations parameter was varied and in which the stochasticity and mutation rate parameters were set to 0.0 and 0.02, respectively (i.e. to the optimal values).

As can be seen, the PSHC algorithm enables the evolving circuits to find optimal solutions in all replications and, in the case of the best parameters (MutRate=0.02 and Stochasticity=0.0, Variations=100), after only 275,300 evaluations. The results obtained with the PSHC algorithm are significantly better than the results obtained with $(\mu + 1)$ and $(1 + \lambda)$ algorithms reported above (Kruskal Wallis test p-value $< 10^{-11}$).

As reported above, experiments carried out with a variation of the algorithm in which the population was divided into 10, 5, or 2 sub-populations formed by 2, 4 or 10 individuals, respectively, instead than 20 sub-populations formed by a single individual, led to much worse results. In these experiments the maximum fitness was achieved in only 84%, 62%, and 54% of the replications, respectively.

|  | MutRate 0.01 | MutRate 0.02 | MutRate 0.03 | MutRate 0.04 | MutRate 0.05 |
|---|---|---|---|---|---|
| Stochasticity 0.0 | *[413,270] (54.51) | **[275,300] (50.12)** | *[361,730] (51.82) | *[692,800] (47.83) | 83.33 (42.88) |
| Stochasticity 0.01 | *[516,270] (60.43) | *[568,800] (56.36) | *[815,800] (52.96) | *[853,730] (50.26) | 80 (46.36) |
| Stochasticity 0.03 | *[537,600] (49.52) | *[798,630] (47.70) | *[809,450] (46.93) | 96.66 (45.20) | 83.33 (43.67) |
| Stochasticity 0.05 | *[1,024,530] (47.95) | *[1,065,420] (54.19) | *[1,160,500] (46.94) | 93.33 (51.29) | 73.33 (48.42) |

Table 6. Performance of circuits evolved with the PSHC method in experiments carried out by varying mutation rate and stochasticity. The number of parents ($\lambda$) was set to 20. The number of variations was set to 100. The first number of each cell indicates the fraction of replications that achieved optimal performance. The experiments in which all replications found optimal solutions are indicated with an asterisk followed by the average number of evaluations that were necessary to find optimal solutions indicated with square brackets. The numbers between parentheses indicate the average size of the evolved circuits.

| Variations |  |
|---|---|
| 1 | *[710'300] |
| 10 | 93.33 |
| 50 | *[764'200] |
| 100 | **[275'300]** |
| 150 | *[454'200] |
| 200 | *[398'410] |
| 500 | *[584'650] |

Table 7. Performance of circuits evolved with the PSHC method in experiments carried out by with different number of Variations. The mutation rate and stochasticity parameters were set to 0.2 and 0.0, respectively. The number of parents ($\lambda$) was set to 20. The first number of each cell indicates the fraction of replications that achieved optimal performance. The experiments in which all replications found optimal solutions are indicated with an asterisk followed by the average number of evaluations that were necessary to find optimal solutions indicated with square brackets. The numbers between parentheses indicate the average size of the evolved circuits.

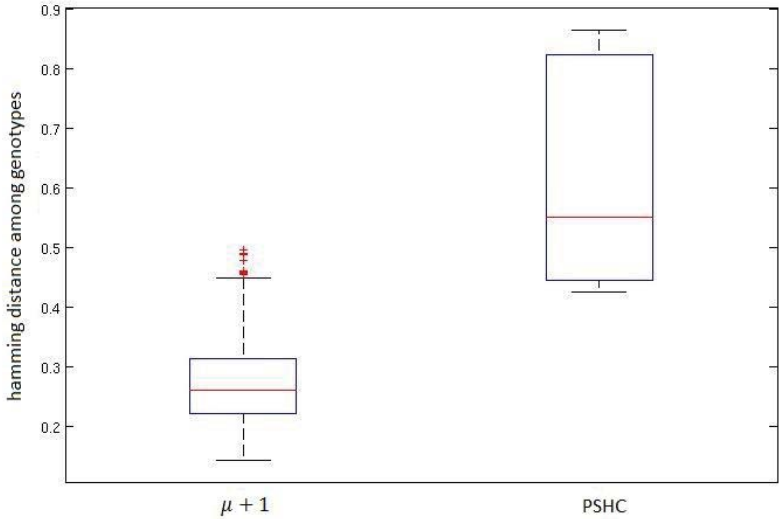

Figure 6. Genetic diversity of the populations evolved with the $(\mu + 1)$ and PSHC algorithms calculated by measuring the average hamming distance between the genotypes of the last generation. Boxes represent the inter-quartile range of the data and horizontal lines inside the boxes mark the median values. The whiskers extend to the most extreme data points within 1.5 times the inter-quartile range from the box. Points marked with the plus sign indicate the outliers.

The comparison of the characteristics of the first circuits that achieved optimal performance evolved with the three different algorithms reveals that circuits evolved with the PSHC have: (i) a greater phenotypic complexity than the circuits evolved with the other two algorithms (Table 8, Kruskal Wallis test p-value < 0.0081), (ii) a greater population diversity than the circuits evolved with the $(\mu + 1)$ and $(1 + \lambda)$ algorithms (Figure 6, Wilcoxon test p-value < $10^{-8}$), and (iii) a greater phenotypic variability than the circuits evolved with the $(\mu + 1)$ algorithm (Table 8, Wilcoxon rank sum test p value < $10^{-5}$) but a lower phenotypic variability with respect to the circuits evolved with the $(\mu + 1)$ algorithm (Table 8, Wilcoxon rank sum test p value < $10^{-7}$)

|  | $(\mu + 1)$ | $(1 + \lambda)$ | PSHC | Kruskal Wallis p-value |
|---|---|---|---|---|
| Phenotypic complexity (number of functional gates) | 61.0 | 61.8 | 71.6333 | 0.0081 |
| Phenotypic variability | 3826.3 | 4826.0 | 4198.5 | 0.0063 |
| Robustness (% offspring with a fitness equal or greater than the parent) | 23.75 | 14.3 | 18.79 | 0.0027 |

Table 8. Comparison of the characteristics of the first circuits evolved with different algorithms that achieved optimal performance (i.e. the 17 out of 30 replications that achieved optimal performance in the case of the $(\mu + 1)$ experiments and the 30 out of 30 replications that achieved optimal performance in the case of the $(1 + \lambda)$ and PSHC experiments).

The comparison between the first circuits that achieved optimal performance (Table 8) and the optimal circuits obtained at the end of the evolutionary process (Table 9) shows how the evolving circuits tend to become progressively more robust with respect to genetic variations during phases in which the fitness remain stable. Indeed the percentage of offspring that preserve the same fitness of their parents increases significantly in all cases (Wilcoxon test p-value < $10^{-15}$). The increased robustness is achieved by reducing the phenotypic complexity and, consequently, by reducing the phenotypic variability.

|                                                                      | $(\mu + 1)$ | $(1 + \lambda)$ | PSHC  | Kruskal Wallis p-value |
|----------------------------------------------------------------------|-------------|-----------------|-------|------------------------|
| Phenotypic complexity (number of functional gates)                   | 48.94       | 44.83           | 43.86 | > 0.05                 |
| Phenotypic variability (individual)                                  | 2617.7      | 2211.1          | 2418.3 | 0.0375                |
| Robustness (% offspring with a fitness equal or greater than the parent) | 33.7    | 27.34           | 39.80 | 0.0058                 |

Table 9. Comparison of the characteristics of the circuits evolved with different algorithms at the end of the evolutionary process. Data calculated by using the first optimal individual circuit of the last generation. In the case of the $(\mu + 1)$ experiments, we included only the replications that achieved optimal performance.

Overall these results indicates that the advantage of the PSHC in this problem domain is due to its ability to limit the effects of the competition between reproducing individuals, that favour the selection of phenotypically simple individuals characterized by low phenotypic variability, and to its ability to maximize the variability of the population.

## 4. Conclusion

In this paper we investigated how the characteristics of the evolutionary algorithm influence the evolvability of candidate solutions, i.e. the propensity of evolving individuals to generate better solutions as a result of random genetic variation. This objective has been pursued by evolving digital circuits which represent a classic domain of application for evolutionary algorithms and which share important properties with natural systems such as proteins, RNA, regulatory circuits and metabolic networks.

    The results indicate that in this domain, which is characterized by a high level of neutrality, $(1 + \lambda)$ ES largely outperform $(\mu + 1)$ ES. The analysis of the evolutionary dynamics indicates that this difference is due to competition for robustness to mutations among evolving individuals. When individuals compete for robustness and when robustness is achieved simply by minimizing the number of genes playing a functional role, as in the case of the circuits evolved with the $(\mu + 1)$ ES, evolution tends to select individuals located in high robust regions of the genetic space which are characterized by low phenotypic variability and consequently low evolvability. When the evolving individuals do not compete for robustness to mutations, as in the case of the circuits evolved with the $(1 + \lambda)$ ES, evolution selects individuals that are less robust but have a higher phenotypic variability and evolvability. Overall these results show how neutrality and robustness to mutations does not necessarily enhance evolvability. They can also reduce evolvability and cause evolutionary stagnation, as also observed by Ancel and Fontana (2000).

    Whether robustness to mutations enhances or diminishes phenotypic variability and evolvability depends on whether robustness is achieved through the development of parsimonious (phenotypically simple) solutions that minimize the number of genes playing functional roles or through phenotypically more complex solutions capable of buffering the effect of mutations. Robustness to mutations of the latter type can evolve as a correlated side effect of the evolution of robustness to environmental variations (De Visser et al., 2003) that cannot be improved through phenotypic simplification. Therefore, whether robustness to mutation enhance or reduce evolvability might depend on whether robustness to mutation is combined or not with robustness to environmental variations. For evidences supporting this hypothesis in the context of evolving digital circuits see Milano and Nolfi (2016).

    The comprehension of the effects that the competition between evolving individuals has on robustness and evolvability enabled us to design a new evolutionary algorithm, named Parallel Stochastic Hill Climber (PSHC), which outperforms the other two methods considered. This is achieved by limiting the negative effects that the competition among evolving individuals can have

on evolvability while preserving the advantage provided by the utilization of a diversified population.

The analysis reported in this paper has been restricted to algorithm operating on the basis of asexual reproduction. The impact of sexual reproduction and of other genetic operators on the evolutionary dynamics should be investigated in future studies.

**References**

Adami C. (2002). Sequence complexity in Darwinian evolution. Complexity, 8:49-57.
Ancel L.W. and Fontana W. (2000). Plasticity, evolvability, and modularity in RNA. Journal of Experimental Zoology part B Molecular Developmental Evolution, 288: 242–283.
Back T. (1994). Selective pressure in evolutionary algorithms: A characterization of selection mechanisms. in Proc. 1st IEEE Conf. Evol. Comput.,Jun. 27–29, 1994, pp. 57–62
Bäck T., and Hammel U. (1994). Evolution strategies applied to perturbed objective functions. In Proceedings of the International Conference on Evolutionary Computation.pp. 40–45.
Balch M. (2003).Complete digital design. New York, NY: McGraw-Hill.
Bedau M.A. and Packard N.H. (2003) Evolution of evolvability via adaptation of mutation rates. Biosystems, 69: 143–162.
Carlson J M, Doyle J (2002). Complexity and robustness. PNAS 99:2538-2545.
Crutchfield J.P., Görnerup O. (2006). Objects that make objects: the population dynamics of structural complexity. Journal of The Royal Society Interface, 3:345-349.
De Visser J. A. et al. (2003). Perspective: Evolution and detection of genetic robustness. Evolution, 57 (9): 1959-1972.
Earl D.J. and Deem M.W. (2004). Evolvability is a selectable trait. PNAS, 101: 11531–11536.
Edelman G.M., and Gally J.A. (2001). Degeneracy and complexity in biological systems. Procedings of the National Academy of Science USA, 98 (13): 763–768.
Hartmann M., and Haddow P. (2004). Evolution of fault tolerant and noise-robust digital designs. Computers and Digital Techniques, IEE Proceedings 151: 287-294.
Hazen RM, Griffin PL, Carothers JM, Szostak JW (2007). Functional information and the emergence of biocomplexity. Proceedings of the National Academy of Sciences, 104:8574-8581.
Houle, D. (1992). Comparing evolvability and variability of quantitative traits. Genetics 130, 195-204.
Hu T., Payne J.L., Banzhaf W, and Moore J.H. (2012). Evolutionary dynamics on multiple scales: a quantitative analysis of the interplay between genotype, phenotype, and fitness in linear genetic programming. Genetic Programming and Evolvable Machines (13) 3: 305–337.
Jin Y. (2005). Evolutionary optimization in uncertain environments: A survey. IEEE Transactions on Evolutionary Computation, 9 (3): 303-317.
Kirschner M. and Gerhart J. (1998). Evolvability. PNAS, 95: 8420–8427.
Koza J. (1992). Genetic Programming: On the Programming of Computers by Means of Natural Selection. Cambridge, MA: MIT Press.
Lenski RE, Barrick JE, Ofria C. (2006). Balancing robustness and evolvability. PLoS Biology, 4: 2190-2192.
Levitan B. and Kauffman S. (1994). Adaptive walks with noisy fitness measurements. Molecular Diversity, (1) 1:53–68.
Macia J., and Solé R.V. (2009). Distributed robustness in cellular networks: Insights from synthetic evolved circuits." Journal of The Royal Society Interface 6 (33): 393–400.
Masel J. and Trotter M.V. (2010). Robustness and evolvability. Trends in Genetics, 26 (9): 406-414.
Milano N., Nolfi S. (2016). Robustness to faults promotes evolvability: Insights from evolving digital circuits, PLoS ONE. 11(7): e0158627
Miller J. and Hartmann M. (2001). Evolving messy gate for fault tolerance: some preliminary findings. In Proceedings 3rd NASA Workshop on Evolvable Hardware, 116-123.

Miller J.F., and Thomson P. (2000). Cartesian Genetic Programming. In Genetic Programming, edited by R. Poli, W. Banzhaf, W. B. Langdon, J. Miller, P. Nordin, and T. C. Fogarty (Eds.) Lecture Notes in Computer Science 1802. Heidelberg: Springer Berlin Heidelberg.

Miller J.F., Job D., and Vassiley V.K. (2000). Principles in the evolutionary design of digital circuits. Journal of Genetic Programming and Evolvable Machines 1 (1): 8-35.

Miller J.F., Thompson A., Thompson P. and Fogarty T. (Eds.) (2000). Proceedings of the 3rd International Conference on Evolvable Systems: From Biology to Hardware. Lecture Notes on Computer Science, no. 1801. Berlin, Germany: Springer Verlag.

Pigliucci M. (2008). Is evolvability evolvable? Nat. Rev. Genet., 9:75–82.

Raman K. and Wagner A. (2011). The evolvability of programmable hardware. Journal of The Royal Society Interface 8 (55): 269–81.

Rana S., Whitlev L.D., and Cogswell R. (1996). Searching in the presence of noise. In H. M. Voigt (Ed.), Parallel Problem Solving from Nature. Lecture Notes in Computer Sciences, 1141:198–207. Berlin: Springer-Verlag.

Schuster P., Fontana W., Stadler P.F. and Hofacker I.L. (1994) From sequences to shapes and back: a case study in RNA secondary structures. Proceedings Royal Society London B 255: 279–284.

Sniegowski P.D. and Murphy H.A. (2006). Evolvability. Current Biology, 16: 831–834.

Thompson A., Layzell P. and Zebulum R. (1999). Explorations in design space: Unconventional electronics design through artificial evolution. IEEE Transactions on Evolutionary Computation 3 (3): 167–96.

Tononi G., Sporns O., and Edelman G.M. (1999). Measures of degeneracy and redundancy in biological networks. Proceedings of the National Academy of Science USA, 96: 3257–3262.

Van Nimwegen E., Crutchfield J. P., Huynen M. (1999). Neutral evolution of mutational robustness. PNAS 96:9716–9720.

Wagner A. (2008). Robustness and evolvability: a paradox resolved. Proceeding of the Royal Society B, 275: 91-100.

Wagner A. (2011). The Origins of Evolutionary Innovations: A Theory of Transformative Change in Living Systems. Oxford, U.K.: Oxford University Press.

Wagner G.P., Altenberg L. (1996). Perspective: Complex adaptations and the evolution of evolvability. Evolution, 50: 967–976.

Whitacre J.M. (2010). Degeneracy: a link between evolvability, robustness and complexity in biological systems. Theoretical Biology and Medical Modelling, 7:6

Whitley D., Rana S., and Heckendorn R.B. (1998). The island model genetic algorithm: On separability, population size and convergence. Journal of Computing and Information Technology, 7:33–47.